\pdfoutput=1

\documentclass[11pt]{article}

\usepackage[final]{acl}

\usepackage{times}
\usepackage{latexsym}

\usepackage[T1]{fontenc}

\usepackage[utf8]{inputenc}

\usepackage{microtype}

\usepackage{inconsolata}

\usepackage{graphicx}

\usepackage{amsmath}
\usepackage{tikz}
\pdfoutput=1
\usepackage[edges]{forest}
\usepackage{xcolor}
\usetikzlibrary{shapes.geometric, arrows}
\definecolor{hidden-black}{RGB}{20,68,106}
\definecolor{purple}{RGB}{144,153,196}
\definecolor{yellow}{RGB}{255,228,123}

\usepackage{booktabs}  
\usepackage{makecell}  
\usepackage{pifont}
\usepackage{enumitem}
\usepackage{float}
\usepackage{amssymb}
\usepackage{multirow}
\usepackage{soul}

\title{A Survey on Sparse Autoencoders: \\Interpreting the Internal Mechanisms of Large Language Models}

\author{Dong Shu\textsuperscript{1,$\dagger$}, Xuansheng Wu\textsuperscript{2,$\dagger$}, Haiyan Zhao\textsuperscript{3,$\dagger$}, Daking Rai\textsuperscript{4}, \\
\textbf{Ziyu Yao\textsuperscript{4}}, \textbf{Ninghao Liu\textsuperscript{2}}, \textbf{Mengnan Du\textsuperscript{3}}\\
\textsuperscript{1}Northwestern University \,
\textsuperscript{2}University of Georgia \,\\
\textsuperscript{3}New Jersey Institute of Technology \,
\textsuperscript{4}George Mason University\\
\small\texttt{dongshu2024@u.northwestern.edu}, \small\texttt{\{xw54582,ninghao.liu\}@uga.edu}, \\ \small\texttt{\{hz54,mengnan.du\}@njit.edu}, \small\texttt{\{drai2,ziyuyao\}@gmu.edu}
}

\begin{document}
\maketitle
\begin{abstract}
Large Language Models (LLMs) have transformed natural language processing, yet their internal mechanisms remain largely opaque. Recently, mechanistic interpretability has attracted significant attention from the research community as a means to understand the inner workings of LLMs. Among various mechanistic interpretability approaches, Sparse Autoencoders (SAEs) have emerged as a promising method due to their ability to disentangle the complex, superimposed features within LLMs into more interpretable components.  This paper presents a comprehensive survey of SAEs for interpreting and understanding the internal workings of LLMs. Our major contributions include: (1) exploring the technical framework of SAEs, covering basic architecture, design improvements, and effective training strategies; (2) examining different approaches to explaining SAE features, categorized into input-based and output-based explanation methods; (3) discussing evaluation methods for assessing SAE performance, covering both structural and functional metrics; and (4) investigating real-world applications of SAEs in understanding and manipulating LLM behaviors.
\end{abstract}

\begin{figure*}
    \centering
    \includegraphics[width=1\linewidth]{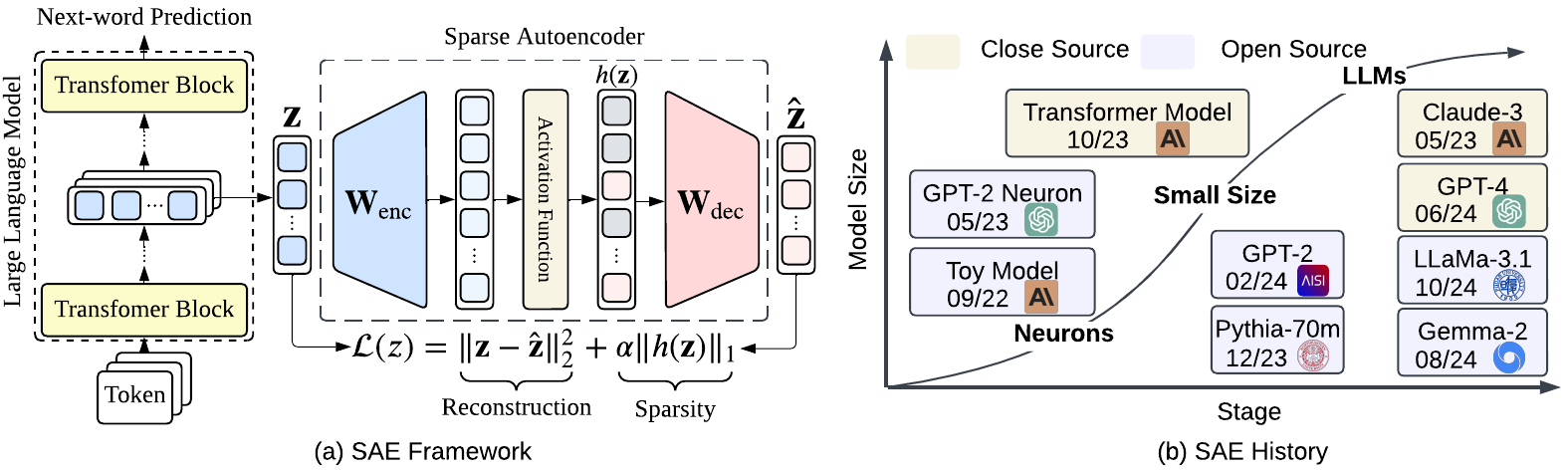}
    \caption{(a) This figure illustrates the fundamental framework of a Sparse Autoencoder (SAE). SAE is trained to take a model representation $\mathbf{z}$ as input and project it to an overcomplete sparse activation $h(\mathbf{z})$ by learning to reconstruct the original input $\hat{\mathbf{z}}$. The SAE typically comprises an encoder, a decoder, and a loss function for training. (b) The development of the SAE progresses through multiple stages. Note that we only list some representative SAE models in this timeline rather than providing an exhaustive compilation.}
    \label{fig:SAE-combined}
\end{figure*}

\section{Introduction}
Large Language Models (LLMs), such as GPT-4 \cite{openai2024gpt4technicalreport}, Claude-3.5 \cite{anthropic2024claude}, DeepSeek-R1 \cite{deepseekai2025deepseekr1incentivizingreasoningcapability}, and Grok-3 \cite{xai2025grok3}, have emerged as powerful tools in natural language processing, demonstrating remarkable capabilities in tasks ranging from text generation to complex reasoning. However, their increasing size and complexity have created significant challenges in understanding their internal representations and decision-making processes. 
This ``black box'' nature of LLMs has sparked a growing interest in mechanistic interpretability~\cite{bereska2024mechanistic, zhao2024explainability, rai2024practical,zhao2024towards}, a field that aims to break down LLMs into understandable components and systematically analyze how these components interact to understand their behaviors. 

Among the various approaches to interpreting LLMs, Sparse Autoencoders (SAEs) ~\citep{cunningham2023sparse, bricken2023monosemanticity, gao2025scaling, rajamanoharan2024jumping, galichin2025have} have emerged as a particularly promising direction for addressing a fundamental challenge in LLM interpretability: \emph{polysemanticity}. Many neurons in LLMs are polysemantic, responding to seemingly unrelated concepts or features simultaneously. This is a phenomenon likely resulting from superposition~\cite{elhage2022superposition}, where LLMs represent more independent features than they have neurons by encoding each feature as a linear combination of neurons. SAEs address this issue by learning an overcomplete, sparse representation of neural activations, effectively disentangling these superimposed features into more interpretable units. By training a sparse autoencoder to reconstruct the activations of a target network layer while enforcing sparsity constraints, SAEs can extract a larger set of \emph{monosemantic} features that offer clearer insights into what information the LLM is processing. This approach has shown promise in transforming the often-inscrutable activations of LLMs into more human-understandable representations, potentially creating a more effective vocabulary for mechanistic analysis of these complex systems.

\subsection{Contribution and Uniqueness}
\noindent\textbf{Our Contributions.}
In this paper, we provide a comprehensive overview of SAE for LLM interpretability, with major contributions listed as following: (1) We explore the technical framework of SAEs, including their basic architecture, various design improvements, and effective training strategies (Section 2). (2) We examine different approaches to analyzing and explaining SAE features, categorized broadly into input-based and output-based explanation methods (Section 3). (3) We discuss evaluation methodologies for assessing SAE performance, covering both structural metrics and functional metrics (Section 4). (4) We discuss real-world applications of SAEs in understanding and manipulating LLMs (Section 5). (5) Additionally, in the appendix, we also introduce key motivations for SAE, discuss connection of SAEs to the broader field of mechanistic interpretability, provide experimental evaluations, and highlight current research challenges and promising future directions.

\vspace{3pt}
\noindent\textbf{Differences with Existing Surveys.}
Several existing surveys take a broad perspective on LLM interpretability. For instance, some surveys provide comprehensive overviews of general explainability methods for LLMs~\cite{ferrando2024primer, zhao2024explainability}, while others focus specifically on mechanistic interpretability as a whole~\cite{rai2024practical,bereskamechanistic}. In contrast, our work uniquely focuses exclusively on SAEs as a specific and promising approach within the mechanistic interpretability landscape. By narrowing our scope to SAEs, we are able to provide a much more comprehensive and detailed analysis of their principles, architectures, training methodologies, evaluation techniques, and practical applications.

\begin{table*}[ht]
    \centering
    \caption{Taxonomy of SAE Frameworks: An Overview of Basic and Variant Architectures.}
    \label{tab:sae_table}
    \scalebox{0.9}{
    \begin{tabular}{l l l l}
    \toprule
    \textbf{Category} & \textbf{Examples} & \textbf{Activation} & \textbf{Citations} \\
    \midrule
    \textbf{Basic SAE Framework (\S\ref{sec:basic_framwork})}
    & $l_2$-norm SAE & ReLU & \citet{ferrando2024primer} \\
    \midrule
    \multirow{6}{*}{\textbf{Improve Architecture (\S\ref{sec:improve_arch})}}
     & Gated SAE & Jump ReLU & \citet{rajamanoharan2024improving} \\
     & TopK SAE & TopK & \citet{gao2025scaling} \\
     & Batch TopK SAE & Batch TopK & \citet{bussmann2024batchtopk} \\
     & ProLU SAE & ProLU & \citet{Taggart} \\
     & JumpReLU SAE & Jump ReLU & \citet{rajamanoharan2024jumping} \\
     & Switch SAE & TopK & \citet{mudide2024efficient} \\
    \midrule
    \multirow{6}{*}{\textbf{Improve Training Strategy (\S\ref{sec:improve_train})}}
     & Layer Group SAE & Jump ReLU & \citet{ghilardi2024efficient} \\
     & Feature Choice SAE & TopK & \citet{ayonrinde2024adaptive} \\
     & Mutual Choice SAE & TopK & \citet{ayonrinde2024adaptive} \\
     & Feature Aligned SAE & TopK & \citet{marks2024enhancing} \\
     & End-to-end SAE & ReLU & \citet{braun2025identifying} \\
     & Formal Languages SAE & ReLU & \citet{menon2024analyzing} \\
     & Specialized SAE & ReLU & \citet{muhamed2024decoding} \\
    \bottomrule
    \end{tabular}}
\end{table*}

\section{Technical Framework of SAEs}
\label{sec:sae_framework}

\subsection{Basic SAE Framework}
\label{sec:basic_framwork}

SAE is a neural network that learns an overcomplete dictionary for representation reconstruction. As shown in Figure \ref{fig:SAE-combined}a, the input of SAE is the representation of a token from LLMs, which is mapped onto a sparse vector of dictionary activations.

\vspace{3pt}
\noindent
\textbf{Input.}\,\, Given a LLM denoted as \( f \) with a total of $L$ transformer layers, we consider an input sequence \( x = (x_0, \dots, x_N) \) with $N$ tokens, where each \( x_n \in x\) represents a token in the sequence. As the sequence \( x \) is processed by the LLM, each token \( x_n \) produces representations at different layers. For a specific layer \( l \), we denote the hidden representation corresponding to token \( x_n \) as \( \mathbf{z}_n^{(l)} \), where \( \mathbf{z}_n^{(l)} \in \mathbb{R}^d \) indicates the embedding vector of dimension \( d \). Each representation \( \mathbf{z}_n^{(l)} \) serves as input to SAEs. In the following, we may omit the superscript $^{(l)}$ of layers to simplify the notation.

After extracting the representation \( \mathbf{z}_n^{(l)} \), the SAE takes it as input, decomposes it into a sparse representation, and then reconstructs it. The SAE framework is typically composed of three key components: the \emph{encoder}, which maps the input representation to a higher-dimensional sparse activation; the \emph{decoder}, which reconstructs the original representation from this sparse activation; and the \emph{loss function}, which ensures accurate reconstruction while enforcing sparsity constraints.

\vspace{3pt}
\noindent
\textbf{Encoding Step.}\,\, Given an input representation \( \mathbf{z} \in \mathbb{R}^{d} \), the encoder applies a linear transformation using a weight matrix \( \mathbf{W}_{\text{enc}} \in \mathbb{R}^{d \times m} \) and a bias term \( \mathbf{b}_{\text{enc}} \in \mathbb{R}^{m} \), 
followed by an activation function $\sigma$ to enforce sparsity. The encoding operation is defined as:
\begin{equation}
h(\mathbf{z}) = \sigma \left( \mathbf{z} \cdot \mathbf{W}_{\text{enc}} + \mathbf{b}_{\text{enc}} \right),
\end{equation}
where \( h(\mathbf{z}) \in \mathbb{R}^{m} \) represents the sparse activation vector, which helps disentangle superposition features. The $\sigma$ activation function could take different formats (see Table~\ref{tab:sae_table}). For example, $\sigma$ could be \( \text{ReLU}\left( x \right) = \max \left( 0, x \right) \), ensures that only non-negative values pass through, encouraging sparsity by setting negative values to zero.

Since the SAE constructs an overcomplete dictionary to facilitate sparse activation, the number of learned dictionary elements \( m \) is chosen to be larger than the input dimension \( d \) (i.e., $m\gg d$). This overcompleteness allows the encoder to learn a richer and more expressive representation of the input, making it possible to reconstruct the original data using only a sparse subset of dictionary elements. The output \( h(\mathbf{z}) \) from the encoder is then passed to the decoding stage, where it is mapped back to the original input space to reconstruct \( \mathbf{z} \).

\vspace{3pt}
\noindent
\textbf{Decoding Step.} After the encoding step, the next stage in the SAE framework is the decoding process, where the sparse activation vector \( h(\mathbf{z}) \) is mapped back to the original input space. This step ensures that the sparse features learned by the encoder contain sufficient information to accurately reconstruct the original representation. The decoding operation is defined as:
\begin{equation}
\hat{\mathbf{z}} = SAE(\mathbf{z}) = h(\mathbf{z})\cdot\mathbf{W}_{\text{dec}} + \mathbf{b}_{\text{dec}},
\end{equation}
where \( \mathbf{W}_{\text{dec}} \in \mathbb{R}^{m \times d} \) is the decoder weight matrix. \( \mathbf{b}_{\text{dec}} \in \mathbb{R}^{d} \) is the decoder bias term. \( \hat{\mathbf{z}} \in \mathbb{R}^{d} \) is the reconstructed output, which aims to approximate the original input \( \mathbf{z} \).

The accuracy of the reconstruction and the interpretability of the learned representation depends heavily on the effectiveness and sparsity of the activation vector \( h(\mathbf{z}) \). Therefore, the SAE is trained using a loss function that balances minimizing the reconstruction error and enforcing sparsity. This trade-off ensures that the learned dictionary elements provide a compact yet expressive representation of the input data.

\vspace{3pt}
\noindent
\textbf{Loss Function.} 
The activation vector \( h(\mathbf{z}) \) is encouraged to be sparse, meaning that most of its values should be zero. Take the ReLU activation for example, while the activation function after the encoder enforces basic sparsity by setting negative values to zero, it does not necessarily eliminate small positive values, which can still contribute to a dense representation. Therefore, additional sparsity enforcement is required. This is achieved using a sparsity regularization term in the loss function, which further promotes a minimal number of active features. Beyond enforcing sparsity, the SAE must also ensure that the learned sparse activation retains sufficient information to accurately reconstruct the original input \( \mathbf{z} \). 
The loss function for training the SAE consists of two key components: \textit{reconstruction loss} and \textit{sparsity regularization}:
\begin{equation}
\mathcal{L}(\mathbf{z}) = \| \mathbf{z} - \hat{\mathbf{z}} \|_2^2 + \alpha \| h(\mathbf{z}) \|_1,
\end{equation}
where reconstruction loss ensures that the SAE learns to reconstruct the input data accurately, meaning the features encoded in the sparse representation must also be present in the input activations. On the other hand, sparsity regularization enforces sparsity by penalizing nonzero values in \( h(\mathbf{z}) \), and $\alpha$ is a hyper-parameter to control the penalty level of the sparsity. Specifically, without the sparsity loss, SAEs could simply memorize the training data, reconstructing the input without disentangling meaningful features. However, once the sparsity loss is introduced, the model is forced to activate only a small subset of neurons for reconstructing the input activation. This constraint encourages the SAE to focus on the most informative and critical features to reconstruct the input activation. A higher value of \( \alpha \) enforces stronger sparsity by shrinking more values in \( h(\mathbf{z}) \) to zero, but this may lead to information loss and degraded reconstruction quality. A lower value of \( \alpha \) prioritizes reconstruction accuracy but may result in less sparsity, reducing the interpretability of the learned features. Thus, selecting an optimal \( \alpha \) is crucial for achieving a balance between interpretability and accurate data representation.





\subsection{Different SAE Variants}
\label{sec:sae_variant}

As SAEs continue to emerge as a powerful tool for interpreting the internal representations of LLMs, researchers have increasingly focused on refining and extending their capabilities. Various SAE variants have been proposed to address the limitations of traditional SAEs, each introducing improvements from different perspectives. In this section, we categorize these advancements into two main groups: Improve Architectural, which modify the structure and design of the traditional SAE, and Improve Training Strategy, which retain the original architecture but introduce novel methods to enhance training efficiency, feature selection, and sparsity enforcement. A taxonomy of representative SAE frameworks is presented in Table \ref{tab:sae_table}. Due to page limitations, examples for each group are provided in Appendix \ref{sec:improve_arch} (Improved Architectural) and Appendix \ref{sec:improve_train} (Improved Training Strategy). We also discuss challenges encountered during SAE training in Appendix \ref{sec:sae_training}.



\begin{figure*}
    \centering
    \includegraphics[width=1\linewidth]{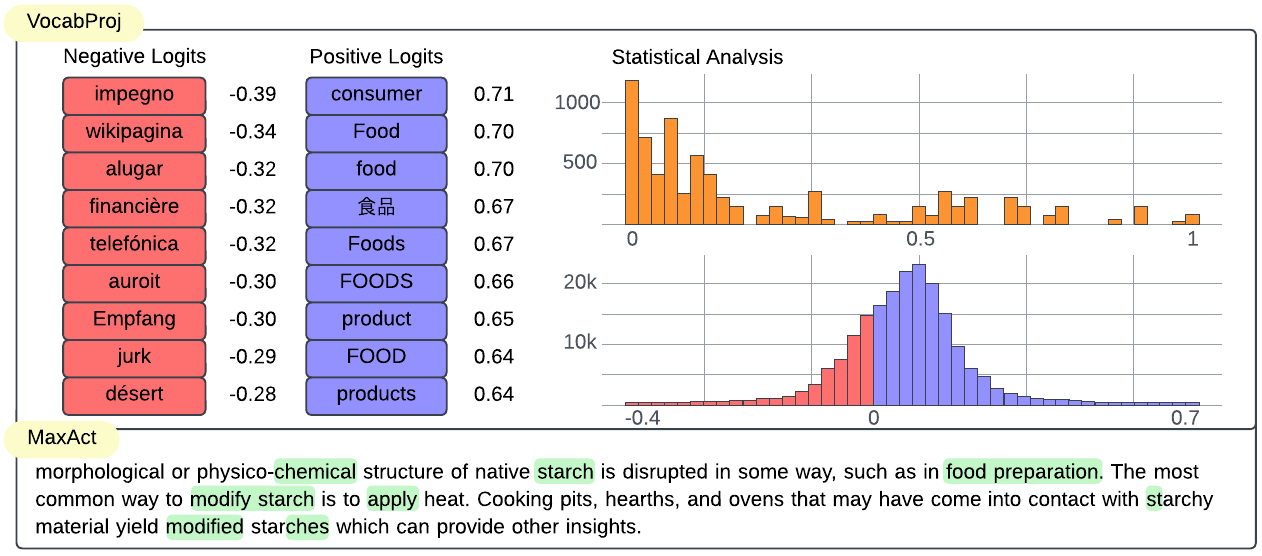}
    \caption{The figure illustrates the interpretation of a learned SAE feature using VocabProj and MaxAct. VocabProj lists words with the highest logits in ``Positive Logits'' column, and lowest logits in ``Negative Logits'' column. The upper histogram in Statistical Analysis shows the distribution of randomly sampled non-zero activations, with the y-axis representing the number of sampled activations and the x-axis indicating activation scores. The lower histogram depicts the logit density, where the y-axis represents the number of tokens and the x-axis corresponds to logit scores. MaxAct highlights tokens in an input text that strongly activate the learned feature. The figure references the Neuronpedia website \cite{neuronpedia}.}
    \label{fig:interpret}
\end{figure*}

\section{Explainability Analysis of SAEs} 
This section aims to interpret the learned feature vectors from a trained SAE with natural language. 
Specifically, given a pre-defined vocabulary set $\mathcal{V}$, the goal of the explainability analysis is to extract a subset of words $\mathcal{I}_m\subset \mathcal{V}$ to represent the meaning of $\mathbf{w}_m=\mathbf{W}_\text{dec}[m]$, for $m=1, ..., M$. 
Humans can understand the meaning of $\mathbf{w}_m$ by reading their natural language explanations $\mathcal{I}_m$. 
There are two lines of work for this purpose, namely the \textit{input-based} and \textit{output-based} methods. 
Figure~\ref{fig:interpret} visualizes generated explanations of using different methods to interpret a learned feature vector.

\subsection{Input-based Explanations}

\noindent
\textbf{MaxAct.}\,\, The most straightforward way to collect natural language explanation is by selecting a set of texts whose hidden representation can \textit{maximally activate} a certain feature vector we are interpreting~\cite{bricken2023monosemanticity, lee2023importance}. Formally, given a large corpus $\mathcal{X}$ where each text span $x\in\mathcal{V}^{N}$ consists of $N$ words, the MaxAct strategy finds $K$ text spans that could maximally activate our interested learned feature vector $\mathbf{w}_m$: 
\begin{equation}
    \mathcal{I}_m = \underset{\mathcal{X}^\prime\subset \mathcal{X},|\mathcal{X}|=K}{\arg\max} \sum_{x\in\mathcal{X}^\prime} f_{<l}(x)\cdot \mathbf{w}_m^\top,
    \label{eq:maxact}
\end{equation}
where $f_{<l}(x)$ indicates generating the hidden representation of input text $x$ at the $l$-th layer, and $l$ is the layer our SAE is trained for. 
This strategy is reasonable for interpreting weight vectors of SAEs because of the sparse nature of SAEs, which indicates that a learned feature vector should only be activated by a certain pattern/concept. 
Therefore, summarizing the most activated text spans can give us a clue to understanding the semantic meaning encoded by the learned feature vector. 

\vspace{3pt}
\noindent
\textbf{PruningMaxAct.}\,\, While MaxAct collects text spans that maximally activate a feature vector, these spans often contain extraneous or redundant phrases that can obscure the underlying concept. Building on the Neuron-to-Graph approach~\cite{foote2023neuron}, researchers~\cite{gao2025scaling} introduce a \textit{pruning} operation to remove irrelevant tokens from each text span, thereby retaining only the minimal context necessary to preserve strong activation. Formally, let $p(\cdot)$ be a pruning strategy that maps text $x$ to $p(x)$, and let $p^{-1}(\cdot)$ recover the original text from its pruned version. The final pruned spans are then gathered via:
\begin{equation}
\begin{aligned}
   &\mathcal{I}_m = \underset{\mathcal{X}^\prime\subset p(\mathcal{X}),\,|\mathcal{X}^\prime|=K}{\arg\max}\; \sum_{x\in\mathcal{X}^\prime} f_{<l}(x)\cdot \mathbf{w}_m^\top, \\
   &\text{s.t. } \forall x\in p(\mathcal{X}), \frac{f_{<l}(p^{-1}(x))\cdot \mathbf{w}_m^\top}{f_{<l}(x)\cdot \mathbf{w}_m^\top}\geq 0.5,   
\end{aligned}
\end{equation}
where the condition enforces that the pruned text $p(x)$ retains at least half of the original activation.
In practice, $p(\cdot)$ can be instantiated by removing selected tokens or replacing them with padding.
According to \citet{gao2025scaling}, this PruningMaxAct technique yields higher recall (i.e., finds more relevant examples) but lower precision compared to the original MaxAct strategy.

\subsection{Output-based Explanations}

\noindent 
\textbf{VocabProj.}\,\, Output-based explanations project the learned feature vectors to the \textit{output word embeddings} of words to compute the activations. 
Mathematically, $f_\text{out}(w):\mathcal{V}\rightarrow\mathbb{R}^d$ denotes the output word embedding layer that returns the output embeddings of a word $w$, and we can collect the natural language explanations by:
\begin{equation}
    \mathcal{I}_m=\underset{\mathcal{V}^\prime\subset \mathcal{V},\,|\mathcal{V}^\prime|=K}{\arg\max}\; \sum_{w\in\mathcal{V}^\prime} f_\text{out}(w)\cdot \mathbf{w}_m^\top.
\end{equation}
This mapping process makes sense for decoder-only LLMs because the layers in such models share the same residual stream, enabling the representations in the intermediate layers to be linear correlated to the output word embeddings~\cite{Nostal2020Logit}. 
Recently, researchers~\cite{wu2025interpreting,gur2025enhancing} find that output-based explanations show a stronger promise in interpreting and controlling LLM behaviors (i.e., generated texts) compared to the input-based ones.

\vspace{3pt}
\noindent
\textbf{MutInfo.}\,\, The VocabProj assumes that the output embeddings that maximally activate an interested feature vector can best describe the meaning of the learned feature. However, this assumption may fail for frequent words, whose embeddings often have large $l_2$ norm~\cite{gao2019representation}. 
To address this, \citet{wu2025interpreting} proposes extracting a vocabulary subset that maximizes mutual information with the learned feature.
Formally, let $\mathcal{C}$ denote knowledge encoded by $\mathbf{w}_c$, the explanations are extracted by 
\begin{equation}
\small
    \begin{aligned}
    \mathcal{I}_m &= \underset{\mathcal{V}^\prime\subset \mathcal{V},|\mathcal{V}^\prime|=M}{\arg\max}\text{MI}(\mathcal{V}^\prime;\mathcal{C}) \propto \underset{\mathcal{V}^\prime\subset \mathcal{V},|\mathcal{V}^\prime|=M} {\arg\min} \text{H}(\mathcal{C}|\mathcal{V}^\prime)  \\[1ex]
    &\propto\underset{\mathcal{V}^\prime\subset \mathcal{V},|\mathcal{V}^\prime|=M}{\arg\max} \sum_{w\in \mathcal{V}^\prime} p(w|\mathbf{w}_m) \log p(\mathbf{w}_m|w),
    \label{eq:original}
\end{aligned}
\end{equation}
where $\text{MI}(\cdot;\cdot)$ indicates mutual information between two variables~\cite{cover1999elements}, $\text{H}(\cdot|\cdot)$ is the conditional entropy, and $U(\mathcal{C})$ includes all possible vectors that express the knowledge $\mathcal{C}$. 
Practically, the conditional probabilities can be estimated by:
\begin{equation}
\small
\begin{aligned}
    p(w|\mathbf{w}_m) &= \frac{\text{exp}(f_\text{out}(w)\cdot \mathbf{w}_m^\top)}{\sum_{w^\prime \in \mathcal{V}}\text{exp}(f_\text{out}(w^\prime)\cdot\mathbf{w}_m^\top)}, \\[1ex]
    p(\mathbf{w}_c|w) &= \frac{\text{exp}(f_\text{out}(w)\cdot\mathbf{w}_c^\top)}{\sum_{c^\prime \in \mathcal{C}}\text{exp}(f_\text{out}(w)\cdot\mathbf{W}_{c^\prime}^\top)}. 
\end{aligned}  
\end{equation}
Compared with VocabProj, that only considers $p(w|\mathbf{w}_m)$, this mutual-information-driven objective highlights the need to normalize the raw activation with $p(\mathbf{w}_m|w)$. 
That is, if a word whose output embedding consistently activates various feature vectors, it loses specification for interpretation.

\section{Evaluation Metrics and Methods} 

Evaluating SAEs is inherently challenging due to the absence of ground truth labels. Unlike traditional machine learning tasks where performance can be directly measured against labeled data, the quality of an SAE must be inferred through a diverse set of metrics. These metrics assess both the internal structure of the model and its functional utility. To provide a comprehensive evaluation framework, we categorize SAE evaluation methods into two main groups: structural metrics and functional metrics. This categorization ensures a holistic assessment of SAEs, covering both their training behavior and real-world applicability.

\subsection{Structural Metrics}

Structural metrics focus on assessing whether an SAE behaves as intended during training. SAEs are designed to optimize both reconstruction fidelity and sparsity, as these properties are explicitly enforced in the training loss. Therefore, natural evaluation metrics assess reconstruction accuracy and sparsity in the model’s learned representations. 
\noindent
\textbf{Reconstruction Fidelity.}\,\,
The most fundamental way to evaluate reconstruction fidelity is through Mean Squared Error (MSE) and Cosine Similarity \cite{ng2011sparse}, which directly compare the original activations with SAE-reconstructed activations. Additional metrics such as Fraction of Variance Unexplained (FVU) (also known as normalized loss) \cite{gao2025scaling} and Explained Variance \cite{karvonensaebench} measure how much variance in the original data is retained after SAE reconstructs. Beyond direct reconstruction comparisons, researchers also evaluate how SAEs affect the probability distribution of model outputs. Cross-Entropy Loss \cite{shannon1948mathematical} and KL Divergence \cite{kullback1951information} measure the shift in probability distributions when substituting original model activations with SAE-generated activations. If the SAE faithfully reconstructs activations, the probability distributions should remain similar. Similarly, Delta LM Loss \cite{lieberum2024gemma} quantifies the difference between the original language model loss and the loss incurred when replacing activations with those from the SAE. Another important aspect of reconstruction fidelity is magnitude preservation. The L2 Ratio \cite{karvonensaebench} compares the Euclidean norms of different activations to ensure that the SAE does not systematically alter activation magnitudes.

\noindent
\textbf{Sparsity.}\,\,
A key design objective of SAEs is sparsity, which ensures that only a small subset of latent neurons activate for any given input. The most direct metric for sparsity is L0 Sparsity \cite{louizos2017learning}, which measures the average number of nonzero activations per input. However, sparsity is not just about minimizing activations; it also requires ensuring that the active features are meaningful. To assess feature usage patterns, Latent Firing Frequency \cite{he2024llama} and Feature Density Statistics \cite{karvonensaebench} track how often each SAE latent is activated across different inputs, ensuring that features are neither too frequent nor inactive. Additionally, the Sparsity-fidelity Trade-off \cite{gao2025scaling} evaluates whether adjusting sparsity affects reconstruction quality, helping to determine the optimal balance between sparsity and fidelity.

\subsection{Functional Metrics}

While structural metrics ensure that an SAE follows its design principles, functional metrics assess whether the SAE is useful for real-world analysis. These include interpretability, which assesses whether the SAE's learned features correspond to meaningful and distinct concepts, and robustness, which evaluates whether the learned representations are stable and generalizable. 

\noindent
\textbf{Interpretability.}\,\,
One of the primary motivations for SAEs is to enhance interpretability by disentangling LLM activations into meaningful features. A crucial property for interpretability is monosemanticity, where each feature should encode a single concept. RAVEL and Automated Interpretability \cite{karvonensaebench} automatically evaluate monosemanticity by using a language model to generate and assess feature descriptions. These methods analyze the most activating contexts for each feature and assign interpretability scores. Sparse Probing and Targeted Probe Perturbation (TPP) \cite{karvonensaebench} evaluate whether SAE features align with specific downstream tasks. In sparse probing, a linear probe is trained using only a small subset of SAE activations, while TPP measures how much perturbing individual latents impacts probe accuracy. If a small number of active features enable strong performance, the SAE has learned disentangled and meaningful representations. Beyond evaluating feature alignment, it is also crucial to assess the faithfulness of feature descriptions. Input-Based Evaluation and Output-Based Evaluation \cite{gur2025enhancing} provide a framework for verifying whether feature descriptions accurately reflect what a feature represents. Input-Based Evaluation tests whether a given feature description correctly identifies which inputs activate the feature by generating activating and neutral examples and measuring activation differences. Output-Based Evaluation assesses whether a feature description captures the causal influence of the feature on model outputs by modifying feature activations and comparing the resulting generated texts. Feature Absorption \cite{karvonensaebench} assesses whether a feature is capturing multiple independent concepts instead of a single interpretable concept. If adding more features does not significantly improve representation quality, it suggests that the extracted features are already sufficient. Another approach to detecting whether each neuron is monosemantic is checking for redundancy or overlap with other neurons. Feature Geometry Analysis \cite{he2024llama, bricken2023monosemanticity, templeton2024scaling} detects redundancy among SAE latents by measuring cosine similarity between decoder columns. If two features have high cosine similarity, they may represent redundant concepts rather than independent units.

\noindent
\textbf{Robustness.}\,\,
In addition to being interpretable, a well-designed SAE should be robust in various contexts. Robustness ensures that SAEs do not overfit to a specific dataset or condition but instead generalize effectively. Generalizability \cite{he2024llama} assesses whether SAEs remain effective when applied to out-of-distribution data. Two common tests for generalizability include evaluating whether SAEs trained on shorter text sequences still perform well on longer sequences and checking whether SAEs trained on base LLM activations generalize to instruction-finetuned models. Unlearning \cite{karvonensaebench} measures whether an SAE can selectively forget specific features while preserving useful information. This is crucial for applications that require privacy-focused models, where sensitive information needs to be erased. Spurious Correlation Removal (SCR) \cite{karvonensaebench} tests whether an SAE can eliminate biased correlations in downstream models. If removing certain latents reduces unwanted correlations without harming performance, the SAE has learned to capture and remove spurious patterns.

Moreover, we provide a comprehensive comparison of SAEs using both structural and functional metrics in Appendix \ref{sec:evaluate_sae}.

\begin{figure*}
    \centering
    \includegraphics[width=1\linewidth]{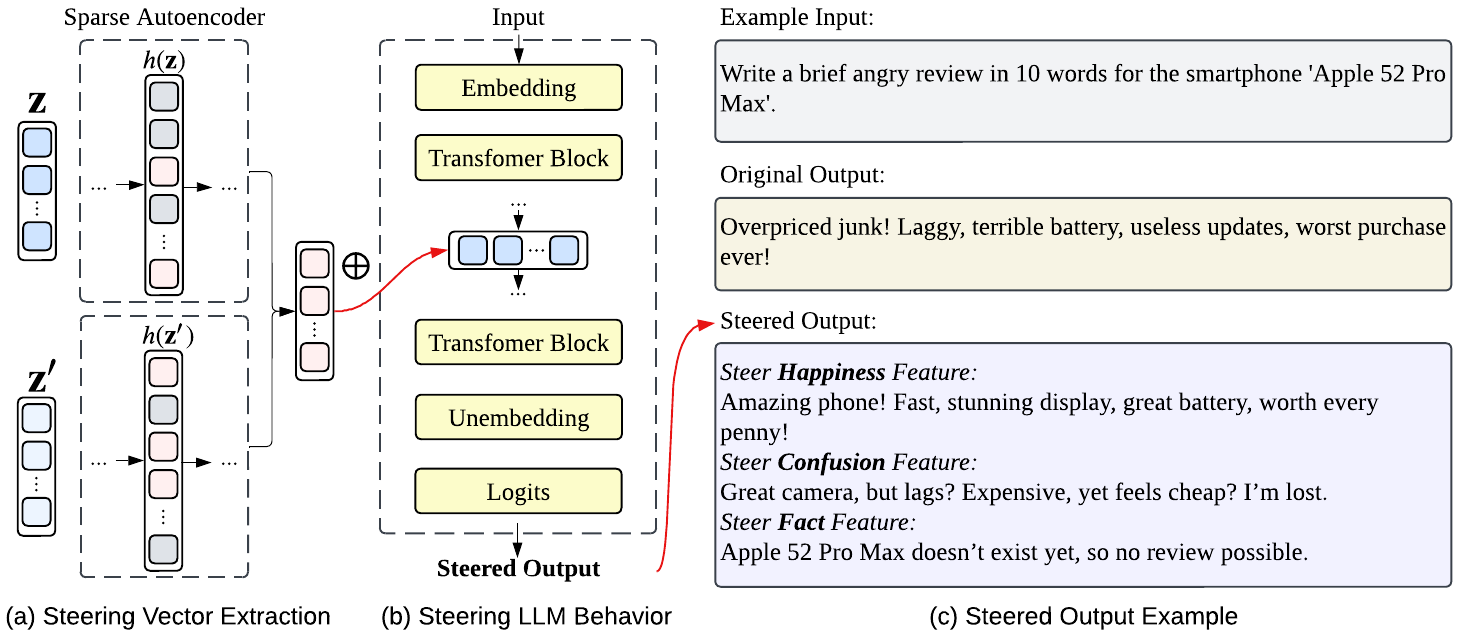}
    \caption{The figure illustrates the process of using a SAE to steer the behavior of a LLM, with an example of the resulting steered output. In part (a), normally people use SAE to extract a steering vector by comparing two representations: $\mathbf{z}$, which lacks a certain feature, and $\mathbf{z'}$, which contains that feature. In part (b), this steering vector is added to the input representation, modifying the LLM’s behavior to align with the desired feature. Part (c) demonstrates the example results of this process, where the steered output reflects the steered feature, even when the original input prompt is neutral or contradictory to the feature being introduced.}
    \label{fig:sae-steer}
\end{figure*}



\section{Applications in Large Language Models}
The latents learned by SAEs represent a collection of low-level concepts. Each SAE latent can be interpreted through gathering its activating examples~\cite{neuronpedia}. This approach enables latents to be interpreted in a human-understandable manner, thereby enhancing our comprehension of how models perform tasks and facilitating more effective control of model behaviors.

\subsection{Model Steering}
Unlike supervised concept vectors, such as probing classifiers~\cite{belinkov-2022-probing,zhao2024beyond,jin2025exploring}, SAEs can simultaneously learn a large volume of concept vectors. The learned vectors can then be utilized to steer model behaviors in ways similar to supervised vectors (See Figure~\ref{fig:sae-steer}). A detailed comparison between SAEs and probing-based methods is provided in Appendix~\ref{sec:sae_probing}.

As previously mentioned, SAE latents can be employed to produce steering vectors that control model outputs related to toxicity, sycophancy, refusal, and emotions. Other steering applications have also proven feasible. Recently, a study shows that SAE latents are able to capture instructions such as translations. These identified latents are effective in manipulating models to translate inputs according to specific instructions~\cite{he2025saif}. Another investigation focusing on semantic search demonstrates that SAEs can learn fine-grained semantic concepts at various levels of abstraction. These concept vectors can be used to steer models toward related semantic~\cite{o2024disentangling}. Alternatively, SAEs trained on biological datasets can provide biology-related features that enable the unlearning of biology-relevant knowledge with fewer side effects than existing techniques~\cite{farrell2024applying}. Given that steering effects are generally challenging to control, SAE-TS further utilizes SAE latents to optimize steering vectors by measuring the changes in SAE feature activation caused by steering, thereby helping construct vectors that target relevant features while minimizing unwanted effects~\cite{chalnev2024improving}. Moreover, explanations based on SAE latents risk prioritizing linguistic features over semantic meaning. \citet{wu2025interpreting} propose a novel approach that promotes diverse semantic explanations, which has been demonstrated to enhance model safety.

\subsection{Model Behavior Analysis}
SAEs construct a dictionary of concepts through their latents, providing a more fine-grained perspective for concept interpretation. This capability enables the analysis of the model's internal representations and learned knowledge in greater detail.

A recent study utilizes SAEs to reveal the mechanism of hallucination, where entity recognition plays a pivot role in recalling facts. A direction distinguishing whether the model knows an entity is identified, which is usually used for hallucination refusal in chat models~\cite{ferrando2025do}. Some studies focus on interpreting how in-context learning (ICL) is performed within models. One study focuses on general ICL tasks, and task-related function vectors were successfully isolated~\cite{lesswrongExtractingTask}. Another study focuses on reinforcement learning (RL) tasks. Their experiment shows that an LLM's internal representations are capable of capturing temporal difference errors and Q-values that are essential in RL computations~\cite{demircan2025sparse}. Besides, one study attempts to examine the working mechanism of instruction following. Their analysis on translation tasks shows that instructions are composed of multiple relevant concepts rather than individual ones~\cite{he2025saif}.

Moreover, SAEs have been employed to study behaviors related to toxicity, sycophancy, refusal, and emotions. One recent study shows that features captured by SAEs can be used to construct probes to classify cross-lingual toxicity~\cite{gallifant2025sparse}. By examining SAE latents that activate on anger-related tokens, researchers have identified steering vectors that control angry outputs~\cite{lesswrongFullPost}. Additionally, other research demonstrates that SAEs can reconstruct vectors responsible for refusing to answer harmful questions as well as directions that produce sycophantic responses~\cite{lesswrongFeaturesRefusal}. Please refer to Appendix \ref{sec:applications_llm} for more LLM applications.

\section{Conclusions}
In this survey, we provided a comprehensive examination of SAEs as a promising approach to interpreting and understanding LLMs. SAEs effectively address the challenge of polysemanticity through learning overcomplete, sparse representations that disentangle superimposed features into more interpretable units. We have systematically explored the foundational principles, technical frameworks, evaluation methodologies, and real-world applications of SAEs in the context of LLM analysis. While SAEs have demonstrated considerable success in revealing the internal mechanisms of LLMs, several challenges remain. In Appendix \ref{sec:research_challenge}, we discussed detailed research challenges, including the incompleteness of concept dictionaries, limited theoretical foundations, persistent reconstruction errors, and substantial computational requirements. Despite these challenges, SAEs continue to evolve through architectural innovations and improved training strategies, offering deeper insights into the inner workings of complex LLMs.

\clearpage
\section*{Limitations}

Many concepts central to SAEs, such as polysemanticity, superposition, and feature disentanglement, have been extensively studied under different frameworks including distributed representations, disentangled representation learning, and sparse coding in computer vision and signal processing. This paper focuses specifically on introducing SAEs during the LLM era, examining their applications in interpreting transformer-based language models, and does not extensively cover these concepts' history in adjacent fields. 



\section*{Acknowledgments}
Mengnan Du is in part supported by National Science Foundation (NSF) Grant \#2310261. 
Ninghao Liu is supported by the National Science Foundation (NSF) Grant \#2223768 and \#2507128. The
views and conclusions in this paper are those of the authors and should not be interpreted as representing any funding agencies.

\bibliography{custom}

\newpage

\clearpage
\appendix

\section{Why Sparse Autoencoders?}
\label{appendix:why_and_history}

As LLMs continue to grow rapidly in size, interpretation becomes more challenging, as the complexity of their latent space and internal representations also expands exponentially.
SAEs have emerged as a powerful tool to understand how LLMs make decisions. This ability is known as mechanistic interpretability, which aims to reverse-engineer models by breaking down their internal computations into understandable, interpretable components. SAE is designed to learn a sparse, linear, and decomposable representation of the internal activations of a LLM. It enforces a sparsity constraint so that only a few features are active at any given time. This encourages each active feature to correspond to a specific, understandable concept. This simplification allows researchers to focus on a few key features rather than being overwhelmed by the full complexity of the model. Below, we discuss the development history of the SAE for LLMs and present Figure \ref{fig:SAE-combined}b to visually depict this progress.
Due to page limitations, we do not attempt to provide an exhaustive history of SAEs, but instead focus on highlighting key milestones in the development of \emph{SAEs for mainstream LLMs.}

\vspace{5pt}
\noindent\textbf{Explaining Individual Neurons.}\,\,

\noindent
The development of interpretability techniques for LLMs has progressed in stages rather than as a single step.
From 2022 to 2023, researchers at OpenAI and Anthropic focused on understanding LLMs by examining individual neurons. OpenAI, for instance, leveraged GPT‑4 to generate natural language explanations for neurons in models like GPT‑2, attempting to map specific neuron activations to concrete linguistic or conceptual features \cite{bills2023language}. Similarly, Anthropic built small toy models trained on synthetic data to observe how features are stored in neurons. These early investigations showed that analyzing single neurons can provide initial insights \cite{elhage2022superposition}. In addition, it is worth noting that the study of explaining individual neurons by labeling interpretable features to them has been extensively explored in studies~\citep{radford2017learning, donnelly2019interpretability, nguyen2019understanding, szegedy2013intriguing} prior to the introduction of mechanistic interpretability.

However, they soon discovered that analyzing individual neurons had significant limitations, as neurons in LLMs often exhibit \emph{polysemanticity}, i.e., responding to multiple unrelated inputs within the same neuron. For instance, a single neuron might simultaneously activate for academic citations, English dialogue, HTTP requests, and Korean text~\cite{bricken2023monosemanticity}. This phenomenon is largely attributed to \emph{superposition}, where neural networks represent more independent features than available neurons by encoding each feature as a linear combination of neurons~\cite{ferrando2024primer}. While this architectural efficiency allows models to encode vast amounts of knowledge, it makes individual neurons difficult to interpret since their activations represent entangled mixtures of different concepts. This fundamental challenge with neuron-level analysis motivated researchers to explore more sophisticated approaches for disentangling these superimposed features, leading to the development of sparse autoencoders (SAEs) as a promising solution for extracting interpretable, monosemantic features from the model's complex internal representations.

\vspace{5pt}
\noindent\textbf{SAEs for Small-size Language Models.}\,\,

\noindent
In late 2023, Anthropic advanced transformer interpretability by moving beyond raw neuron activations to decompose model activations into single-concept, monosemantic features, addressing the polysemanticity of individual neurons in LLMs \cite{cunningham2023sparse, bricken2023monosemanticity}. They trained SAEs on transformer activation data by optimizing a reconstruction loss with a strong sparsity constraint. This training forces the autoencoder to represent each activation as a sparse combination of basis vectors, with each basis vector ideally capturing a distinct, interpretable concept. SAEs transform the overlapping signals of individual neurons into a set of clean, monosemantic features that are much easier to understand. This approach offers a clear advantage over traditional neuron-based methods by isolating the key features that drive model behavior. The promising experimental results on simpler transformer models demonstrated that SAEs provide a more effective and scalable route for interpreting model internals. Building on this success, later works \citet{bloom2024gpt2residualsaes} and \citet{marks2024dictionary_learning} applied SAE techniques to smaller models such as GPT-2~\citep{radford2019language} and Pythia-70m~\citep{biderman2023pythia}, thereby paving the way for their eventual extension to full-scale billion size LLMs.

\vspace{5pt}
\noindent\textbf{SAEs for Large Language Models.}\,\,

\noindent
After witnessing the success of SAEs on smaller-scale models, the third stage of their development emerged in 2024, when Anthropic \cite{templeton2024scaling} and OpenAI \cite{gao2025scaling} became the first groups to apply SAEs to their latest proprietary LLMs, Claude 3 Sonnet and GPT-4, respectively. This marked a significant step forward in understanding these closed-source, black-box models, even for the researchers who built them. However, scaling SAEs from small models to full-scale LLMs introduced several new challenges. One major issue was the sheer scale of activations in models with billions of parameters, which made training and extracting interpretable features computationally expensive. Additionally, ensuring that extracted features remained monosemantic became increasingly difficult, as feature superposition is more prevalent in larger models \cite{templeton2024scaling}. Despite these challenges, researchers found that SAEs could effectively decompose polysemantic neurons into monosemantic features, revealing meaningful and interpretable latent representations within the models. For instance, Anthropic demonstrated that certain neurons in Claude 3 Sonnet encode high-level concepts such as ``sycophantic praise'', where phrases like ``a generous and gracious man'' strongly activate this feature. Similarly, OpenAI's research on GPT-4 identified a ``humans have flaws'' feature, which activates on phrases like ``My Dad wasn’t perfect (are any of us?) but he loved us dearly.'' These findings not only deepen our understanding of model behavior but also provide powerful interpretability tools, allowing the practitioners to better analyze, refine, and steer language model outputs.

As the architecture and mechanisms of SAEs become clearer, more researchers have begun to follow this approach, applying SAEs to interpret open-source models. For example, Google DeepMind \cite{lieberum2024gemma} used SAEs to analyze Gemma 2~\citep{team2024gemma}, while \citet{he2024llama} applied similar techniques to LLaMA 3.1~\citep{dubey2024llama}. This growing adoption highlights the increasing role of SAEs in mechanistic interpretability, paving the way for broader transparency in both close- and open-source LLMs.

\section{Connection of SAEs to the Broader Field of Interpretability}
The field of mechanistic interpretability (MI) has been critiqued for its insufficient engagement with the broader interpretability and NLP research literature~\cite{bereska2024mechanistic, saphra2024mechanistic}. Many of the research topics within MI, such as polysemanticity, superposition, and SAEs, have been investigated in prior and concurrent non-MI fields, often under different terminologies while addressing the same fundamental challenges~\citep{saphra2024mechanistic, elhage2022superposition}. For instance, the study of polysemanticity and superposition, which aims to understand how features are encoded in the model activations, have been studied in the context of distributed representations~\citep{hinton1984distributed, mikolov2013linguistic, mikolov2013distributed, arora2018linear, olah2023distributed}, disentangled representations~\citep{higgins2018towards, kim2018disentangling, locatello2019challenging}, and concept-based interpretability~\citep{nicolson2024explaining, kim2018interpretability}. Similarly, SAEs are closely related to and draw inspiration from earlier lines of research on sparse coding and dictionary learning~\citep{olshausen1997sparse, gregor2010learning, faruqui2015sparse, subramanian2018spine}. These methods, like SAEs, posit the feature sparsity hypothesis~\citep{elhage2022superposition} and aim to learn an overcomplete representation to disentangle the features from activation in superposition. Since these fields pursue similar goals or study the same research problems, the current disconnect causes issues such as missing relevant literature, hindering collaboration, unintentionally redefining established concepts, rediscovering existing techniques, and overlooking well-known baselines. Therefore, it is imperative for the MI community to bridge these gaps and more actively integrate findings from related non-MI research.

\section{Different SAE Variants}

\subsection{Improve Architecture}
\label{sec:improve_arch}

\noindent
\noindent
\textbf{Gated SAE.}\,\, The Gated SAE \cite{rajamanoharan2024improving} is a modification of the standard SAE that aims to improve the trade-off between reconstruction fidelity and sparsity enforcement. Traditional SAEs suffer from shrinkage bias \cite{Wright_Sharkey}, where the \( L_1 \)-norm regularization systematically underestimates feature activations, leading to reduced reconstruction accuracy. Inspired by Gated Linear Units \cite{dauphin2017language, shazeer2020glu}, Gated SAEs replace the standard ReLU encoder with a gated ReLU encoder, which separates the roles of detecting which features are active and estimating their magnitudes. Compared to traditional SAEs, this architectural separation addresses the shrinkage effect caused by L1 regularization, which adds a linear penalty to all nonzero activations, leading to systematic underestimation of feature magnitudes.
\begin{equation}
\tilde{h}(\mathbf{z}) = \mathbf{1} \left[ \pi_{\text{gate}}(\mathbf{z}) > 0 \right] \odot \text{ReLU}(\pi_{\text{mag}}(\mathbf{z})),
\end{equation}
where \( \pi_{\text{gate}}(\mathbf{z}) = \mathbf{W}_{\text{gate}}(\mathbf{z} - \mathbf{b}_{\text{dec}}) + \mathbf{b}_{\text{gate}} \) is the gating function that determines which features should be activated. \( \mathbf{W}_{\text{gate}} \) is a weight matrix for feature selection. \( \pi_{\text{mag}}(\mathbf{z}) = \mathbf{W}_{\text{mag}}(\mathbf{z} - \mathbf{b}_{\text{dec}}) + \mathbf{b}_{\text{mag}} \) is the magnitude estimation function that determines the strength of the active features. \( \mathbf{W}_{\text{mag}} \) is a weight matrix for magnitude estimation. \( \mathbf{1}[\cdot] \) is the Heaviside step function that binarizes activations and \( \odot \) denotes element-wise multiplication. In this case, Gated SAEs introduce independent pathways for determining which features are activated and their respective strengths, reducing bias and improving interpretability. 


To optimize the Gated SAE, the authors introduce an auxiliary loss \( \| \mathbf{z} - \hat{\mathbf{z}}_{\text{frozen}} \left( \text{ReLU} \left( \pi_{\text{gate}}(\mathbf{z}) \right) \right) \|_2^2 \) on top of the traditional loss function. This encourages the gating path to produce useful feature selections for reconstruction, without affecting the learned decoder weights. Here, \( \hat{\mathbf{z}}_{\text{frozen}} \) is a copy of the decoder with frozen weights.

\vspace{3pt}
\noindent 
\textbf{TopK SAE.}\,\, The TopK SAE \cite{gao2025scaling} is an improvement over the traditional SAE, designed to directly enforce sparsity without requiring \( L_1 \)-norm regularization. Instead of penalizing all activations, which can introduce shrinkage bias, TopK SAEs enforce sparsity by retaining only the top \( K \) largest activations and setting the rest to zero. This ensures that only the most important features contribute to the learned representation. The encoder applies a linear transformation followed by a hard TopK selection:
\begin{equation}
\tilde{h}(\mathbf{z}) = \text{TopK}\left( \mathbf{W}_{\text{enc}}(\mathbf{z} - \mathbf{b}_{\text{pre}}) \right),
\end{equation}
where \( \mathbf{W}_{\text{enc}} \in \mathbb{R}^{d \times m} \) is the encoder weight matrix, and \( \mathbf{b}_{\text{pre}} \in \mathbb{R}^{m} \) is a pre-normalization bias term applied before the TopK selection.

Since the sparsity constraint is explicitly enforced through the TopK operation, there is no need for an additional sparsity regularization term in the loss function. The training objective reduces to minimizing the reconstruction loss:
\begin{equation}
\mathcal{L}(\mathbf{z}) = \| \mathbf{z} - \hat{\mathbf{z}} \|_2^2 + \alpha \mathcal{L}_{\text{aux}},
\end{equation}
where \( \mathcal{L}_{\text{aux}} \) is an auxiliary loss scaled by the coefficient \( \alpha \), designed to stabilize training and prevent dead latents \cite{templeton2024scaling}.

\vspace{3pt}
\noindent 
\textbf{BatchTopK SAE.}\,\, The BatchTopK SAE is a modification of the TopK SAE, designed to address the limitations of enforcing a fixed number of active features per sample. \citet{bussmann2024batchtopk} identified two key inefficiencies in the standard TopK SAE. First, it forces every token to use exactly \( K \) features, even when some tokens may require fewer or more active features. It also does not allow flexibility across a batch, leading to inefficient sparsity control. To overcome these issues, BatchTopK SAEs apply the TopK selection globally across the entire batch instead of enforcing it per token. This means that BatchTopK selects the top \( K \times B \) activations across the entire batch, where \( B \) is the batch size. The encoder is modified to:
\begin{equation}
\tilde{h}(\mathbf{Z}) = \text{BatchTopK}\left( \mathbf{W}_{\text{enc}}(\mathbf{Z} - \mathbf{b}_{\text{pre}}) \right),
\end{equation}
where \( \mathbf{Z} \in \mathbb{R}^{B \times d} \) is the input batch matrix, and \( B \) is the batch size. Similar to TopK SAE, BatchTopK directly controls sparsity through the selection mechanism, it eliminates the need for explicit sparsity regularization:
\(
\mathcal{L}(\mathbf{Z}) = \| \mathbf{Z} - \hat{\mathbf{Z}} \|_2^2 + \alpha \mathcal{L}_{\text{aux}}
\).

\vspace{3pt}
\noindent 
\textbf{ProLU SAE.}\,\, The ProLU SAE \cite{Taggart} introduces a novel activation function called Proportional ReLU, which serves as an alternative to ReLU in traditional SAEs. ProLU SAE provides a Pareto improvement over both standard SAEs with \( L_1 \)-norm regularization, which suffer from shrinkage bias, and SAEs trained with a Sqrt(\( L_1 \)) penalty, which attempt to mitigate shrinkage but still do not fully address inconsistencies in activation scaling. In contrast to ReLU, which applies a fixed threshold at zero, ProLU introduces a learnable threshold for each activation, allowing the model to determine the optimal activation boundary dynamically. The ProLU activation function is defined as:
\begin{equation}
\small
\text{ProLU}(m_i, b_i) =
\begin{cases}
m_i, & \text{if } m_i + b_i > 0 \text{ and } m_i > 0 \\
0, & \text{otherwise}
\end{cases},
\end{equation}
where \( m_i \) is the pre-activation output from the encoder, and \( b_i \) is a learnable bias term that shifts the activation threshold. The encoding process in ProLU SAE replaces the standard ReLU activation with ProLU, leading to the following encoding function:
\begin{equation}
\tilde{h}(\mathbf{z}) = \text{ProLU}((\mathbf{z} - \mathbf{b}_{\text{dec}}) \mathbf{W}_{\text{enc}}, \mathbf{b}_{\text{enc}}).
\end{equation}

The ProLU SAE training objective consists of the standard reconstruction loss combined with an auxiliary sparsity term:
\begin{equation}
\mathcal{L}(\mathbf{z}) = \| \mathbf{z} - \hat{\mathbf{z}} \|_2^2 + \lambda P(\tilde{h}(\mathbf{z})),
\end{equation}
where $\lambda$ is the sparsity penalty coefficient, and \( P(\tilde{h}(\mathbf{z})) \) is a sparsity-inducing function. The authors found that using a Sqrt(\( L_1 \)) penalty, defined as \( P(h) = \| h \|_{1/2} \), provided better sparsity control compared to the standard \( L_1 \)-norm.

\vspace{3pt}
\noindent 
\textbf{JumpReLU SAE.}\,\, The JumpReLU SAE \cite{rajamanoharan2024jumping} is a modification of the traditional SAE that replaces the standard ReLU activation function with JumpReLU. The ReLU activation function sets all negative pre-activation values to zero but allows small positive values, leading to false positives in feature selection and underestimation of feature magnitudes. JumpReLU introduces an explicit threshold \( \theta \) that zeroes out pre-activations below this threshold, ensuring that weak activations do not contribute to the reconstruction. The JumpReLU activation function is defined as:
\begin{equation}
\text{JumpReLU}_\theta (\mathbf{z}) = \mathbf{z} \cdot H(\mathbf{z} - \theta),
\end{equation}
where \( \theta \) is a learnable threshold and \( H(x) \) is the Heaviside step function, which is 1 when \( x > 0 \) and 0 otherwise. The encoder in JumpReLU SAE follows a standard linear transformation followed by JumpReLU activation:
\begin{equation}
\tilde{h}(\mathbf{z}) = \text{JumpReLU}_\theta (\mathbf{W}_{\text{enc}} \mathbf{z} + \mathbf{b}_{\text{enc}}).
\end{equation}
Unlike traditional SAEs that use \( L_1 \)-norm for sparsity regularization, JumpReLU SAEs directly optimize the \( L_0 \)-norm, which counts the number of nonzero activations:
\(
\mathcal{L}(\mathbf{z}) = \| \mathbf{z} - \hat{\mathbf{z}} \|_2^2 + \alpha \| h(\mathbf{\mathbf{z}}) \|_0
\).

\vspace{3pt}
\noindent 
\textbf{Switch SAE.}\,\, Inspired by Mixture of Experts (MoE) models \cite{shazeer2017outrageously}, Switch SAE \cite{mudide2024efficient} introduces a more computationally efficient framework for training SAEs. Instead of training a single large SAE, Switch SAE leverages multiple smaller ``expert SAEs'' \( E_1, E_2, ..., E_N \) and a routing network that dynamically assigns each input to an appropriate expert. This approach enables efficient scaling to a large number of features while avoiding the memory and FLOP bottlenecks of traditional SAEs. Each ``expert SAE'' follows a standard TopK SAE formulation:
\begin{equation}
E_i(\mathbf{z}) = \mathbf{W}_{\text{dec}}^{(i)} \cdot \text{TopK}(\mathbf{W}_{\text{enc}}^{(i)} \mathbf{z}),
\end{equation}
where \( \mathbf{W}_{\text{enc}}^{(i)} \) and \( \mathbf{W}_{\text{dec}}^{(i)} \) are the encoder and decoder weight matrices for expert \( i \). The routing network determines which expert is assigned to each input by computing a probability distribution over the experts:
\begin{equation}
p(\mathbf{z}) = \text{softmax}(\mathbf{W}_{\text{router}}(\mathbf{z} - \mathbf{b}_{\text{router}})),
\end{equation}
where \( \mathbf{W}_{\text{router}} \) is the routing weight matrix. \( \mathbf{b}_{\text{router}} \) is the bias term. \( p(\mathbf{z}) \) represents the probability of selecting each expert. The final reconstruction is computed as:
\begin{equation}
\hat{\mathbf{z}} = p_{i^*(\mathbf{z})} E_{i^*(\mathbf{z})}(\mathbf{z} - \mathbf{b}_{\text{pre}}) + \mathbf{b}_{\text{pre}},
\end{equation}
where \( i^*(\mathbf{z}) \) is the selected expert for input \( \mathbf{z} \).

To ensure balanced expert utilization and avoid expert collapse, Switch SAE incorporates an auxiliary loss for load balancing:
\begin{equation}
\mathcal{L}_{\text{aux}} = N \sum_{i=1}^{N} f_i \cdot P_i,
\end{equation}
where \( f_i \) is the fraction of activations assigned to expert \( i \), and \( P_i \) is the fraction of router probability assigned to expert \( i \). This auxiliary loss is then added to the traditional reconstruction loss function to form the final learning objective.

\subsection{Improve Training Strategy}
\label{sec:improve_train}

\noindent
\textbf{Layer Group SAE.}\,\, Traditionally, one SAE is trained per layer in a transformer-based LLM, resulting in a substantial number of parameters and high computational costs. To address this inefficiency, the Layer Group SAE \cite{ghilardi2024efficient} clusters multiple layers into groups based on activation similarity and trains a single SAE per group. This significantly reduces training time while preserving reconstruction accuracy and interpretability. To determine which layers should be grouped together, the method measures the angular similarity between layer activations, defined as:
\begin{equation}
\small
d_{\text{angular}}(\mathbf{z}^p_{\text{post}}, \mathbf{z}^q_{\text{post}}) = \frac{1}{\pi} \arccos \left( \frac{\mathbf{z}^p_{\text{post}} \cdot \mathbf{z}^q_{\text{post}}}{\|\mathbf{z}^p_{\text{post}}\|_2 \|\mathbf{z}^q_{\text{post}}\|_2} \right),
\end{equation}
where \( \mathbf{z}^p_{\text{post}} \) and \( \mathbf{z}^q_{\text{post}} \) represent post-MLP residual stream activations for layers \( p \) and \( q \). Using this similarity metric, layers with highly correlated activations are clustered together through a hierarchical clustering strategy. The number of groups \( K \) is chosen based on a computational trade-off, balancing efficiency and reconstruction accuracy. Once the layer groups are formed, a single SAE is trained per group instead of one per layer. The SAE architecture and training objective remains similar as in traditional SAEs, optimizing for both reconstruction accuracy and sparsity.

\vspace{3pt}
\noindent 
\textbf{Feature Choice SAE.}\,\, Traditional SAEs face several limitations, including dead features, fixed sparsity per token, and lack of adaptive computation. Feature Choice SAEs \cite{ayonrinde2024adaptive} address these issues by imposing a constraint on the number of tokens each feature can be active for, rather than restricting the number of active features per token. This approach ensures that all features are utilized efficiently, preventing feature collapse and improving reconstruction accuracy. This sparsity allocation constraint is defined as:
\begin{equation}
\sum_{j} S_{i,j} = m, \forall i, \text{where } M = mF,
\end{equation}
where \( S_{i,j} \) is a binary selection matrix, indicating whether feature \( i \) is active for token \( j \). Each feature must be activated for exactly \( m \) tokens, enforcing uniform feature utilization.

\vspace{3pt}
\noindent 
\textbf{Mutual Choice SAE.}\,\, Mutual Choice SAE \cite{ayonrinde2024adaptive} remove all constraints on sparsity allocation, allowing the model to freely distribute its limited total sparsity budget across all tokens and features. Unlike TopK SAEs, which enforce a fixed number of active features per token, or Feature Choice SAEs, which constrain the number of tokens each feature can be assigned to, Mutual Choice SAE introduce global sparsity allocation. This means that instead of enforcing a per-token or per-feature selection, the model selects the top $M$ feature-token matches across the entire dataset, ensuring that sparsity is allocated adaptively based on reconstruction needs. Mathematically, the activation selection process is defined as:
\begin{equation}
S = \text{TopKIndices}(\mathbf{Z}', M),
\end{equation}
where \( \mathbf{Z}' \) represents the pre-activation affinity matrix between tokens and features. \( M \) is the global sparsity budget, denoting the total number of active feature-token pairs allowed. \( \text{TopKIndices}(\cdot) \) selects the top \( M \) activations globally, instead of enforcing a fixed \( K \) per token.

\vspace{3pt}
\noindent 
\textbf{Feature Aligned SAE.}\,\, The Feature Aligned SAE \cite{marks2024enhancing} introduces Mutual Feature Regularization (MFR), a novel training method designed to improve the interpretability and fidelity of learned features in SAEs. Traditional SAEs often suffer from feature fragmentation, where meaningful input features get split across multiple decoder weights, and feature entanglement, where multiple independent input features are merged into a single decoder weight. These issues reduce the interpretability of SAEs and limit their effectiveness in analyzing neural activations. The key insight behind MFR is that features learned by multiple SAEs trained on the same dataset are more likely to align with the true underlying structure of the input data. To enforce this, Feature Aligned SAE trains multiple SAEs in parallel and applies a MFR penalty that encourages them to learn mutually consistent features:
\begin{equation}
\scriptsize
\mathcal{L}_{\text{MFR}} = \alpha \left(\frac{1}{N(N-1)} \sum_{i=1}^{N-1} \sum_{j=i+1}^{N} (1 - \text{MMCS}(\mathbf{W}_{dec}^{(i)}, \mathbf{W}_{dec}^{(j)})) \right),
\end{equation}
where \( \mathbf{W}_{dec}^{(i)} \) and \( \mathbf{W}_{dec}^{(j)} \) are the decoder weight matrices of different SAEs. Mean of Max Cosine Similarity (MMCS) measures the degree of alignment between the learned features across SAEs. \( \alpha \) is a hyperparameter that controls the strength of the regularization. This mutual feature regularization is then combined with the traditional SAE loss to form the final training objective of the Feature Aligned SAE.

\vspace{3pt}
\noindent 
\textbf{End-to-end SAE.}\,\, Traditional SAEs often prioritize minimizing reconstruction error rather than ensuring that learned features are functionally important to the model’s decision-making. This often leads to feature splitting, where a single meaningful feature is divided into multiple redundant components. To address this, End-to-end SAE \cite{braun2025identifying} modifies the training objective to ensure that the discovered features directly influence the network’s output. They propose minimizing the Kullback-Leibler (KL) divergence between the original network's output distribution and the output distribution when using SAE activations, formulated as:
\begin{equation}
\mathcal{L}_{\text{e2e}} = KL(\hat{y}, y) + \alpha \| h(\mathbf{z}) \|_1.
\end{equation}
To further ensure that activations follow similar computational pathways in later layers, they propose E2e + Downstream SAE, which introduces an additional downstream reconstruction loss, leading to the formulation:
\begin{equation}
\scriptsize
\mathcal{L}_{\text{e2e+ds}} = KL(\hat{y}, y) + \alpha \| h(\mathbf{z}) \|_1 + \beta \sum_{k=l+1}^{L} \| \hat{\mathbf{a}}^{(k)} - \mathbf{a}^{(k)} \|_2^2.
\end{equation}
By shifting the training focus from activation reconstruction to output distribution preservation, this method ensures that learned features are more aligned with the actual computational processes of the network while maintaining interpretability.

\vspace{3pt}
\noindent 
\textbf{Formal Languages SAE.}\,\, Traditional SAEs effectiveness remains questionable in language models due to their reliance on correlations rather than causal attributions. While SAEs often recover features that correlate with linguistic structures, such as parts of speech or syntactic depth, interventions on these features frequently do not influence the model’s predictions, suggesting that current training objectives fail to ensure causal relevance. To address this, Formal Languages SAE \cite{menon2024analyzing} introduce a causal loss term that explicitly encourages SAEs to learn features that impact the model’s computation. Their proposed loss function is given by:
\begin{equation}
L = L_{\text{recon}} + \alpha L_{\text{sparse}} + \beta L_{\text{caus}},
\end{equation}
where \( L_{\text{recon}} \) is the standard reconstruction loss, \( L_{\text{sparse}} \) enforces sparsity, and \( L_{\text{caus}} \) ensures that interventions on learned features result in predictable changes in model output.

\vspace{3pt}
\noindent 
\textbf{Specialized SAE.}\,\, Traditional SAEs struggle to capture rare and low-frequency concepts, which are critical for understanding model behavior in specific subdomains. To address this, Specialized SAE (SSAE) \cite{muhamed2024decoding} focuses on learning rare subdomain-specific features through targeted data selection and a novel training objective. Instead of training on the full dataset, SSAE uses high-recall dense retrieval methods, such as BM25, Contriever, and TracIn reranking, to identify relevant subdomain data, ensuring that rare features are well-represented. Additionally, they introduce Tilted Empirical Risk Minimization (TERM), an objective that optimizes for worst-case reconstruction loss rather than average loss. This is achieved by modifying the standard SAE loss function to:
\begin{equation}
L_{\text{TERM}}(t; w) = \frac{1}{t} \log \left( \frac{1}{N} \sum_{i=1}^{N} e^{t \cdot L_w(\mathbf{z_i})} \right),
\end{equation}
where $L_w(\mathbf{z_i})$ is the standard SAE loss for representation $\mathbf{z_i}$. $N$ is the size of a minibatch, and \( t \) is the tilt parameter that controls emphasis on rare concept reconstruction. 

\subsection{SAE Training}
\label{sec:sae_training}

Even though the framework of SAEs is conceptually straightforward, training SAEs is both computationally expensive and data-intensive. The complexity arises due to the overcomplete dictionary representation, large-scale data requirements, and the layer-wise training paradigm necessary for interpreting LLMs. Each of these factors contributes to the substantial computational cost associated with training SAEs at scale.

\noindent
\textbf{Overcomplete Dictionary Representation.}  
A defining characteristic of SAEs is their overcomplete dictionary, where the number of learned features far exceeds the dimensionality of the LLM's latent space. This overcompleteness is what enables SAEs to enforce sparsity, allowing them to isolate and extract meaningful feature activations from high-dimensional representations. The enforced sparsity is crucial for LLM interpretability, as it helps decompose complex neural activations into more semantically meaningful features. Empirical studies highlight the scale of overcompleteness; for example, LLaMa-Scope \cite{he2024llama} trained SAEs with 32K and 128K features, which are 8× and 32× larger than the hidden size of LLaMa3.1-8B. This extreme overparameterization provides a highly expressive feature space but significantly increases the computational burden during training.

\noindent
\textbf{Large-Scale Data Requirements.}  
Since the input to an SAE consists of representations from LLMs, an enormous amount of data is required to ensure that the model learns a diverse and representative set of activations. To effectively train an SAE, it is essential to activate a wide range of neurons in the LLM, which necessitates processing large-scale datasets covering diverse linguistic structures. Moreover, because SAEs are overcomplete, they require significantly more training data to converge. Empirical results from Gemma-Scope \cite{lieberum2024gemma} illustrate this requirement: SAEs with 16.4K features were trained on 4 billion tokens, while 1M-feature SAEs required 16 billion tokens to reach satisfactory performance. This highlights the immense data demands necessary for training effective SAEs. Another challenge arises when scaling up the training data, which is how to efficiently shuffling massive datasets across distributed systems. Shuffling is crucial to prevent models from learning spurious, order-dependent patterns. However, as datasets grow to terabyte or petabyte scales, performing a distributed shuffle becomes a significant engineering hurdle \cite{challenges}.

\noindent
\textbf{Layer-Wise Training.}  
Interpreting an LLM requires understanding its representations at each layer, which necessitates training separate SAEs for different layers of the model. The standard approach is to train one SAE per layer, meaning that for deep models, this process must be repeated across dozens or even hundreds of layers, compounding the computational cost. The necessity of layer-wise training is further evidenced by ongoing research efforts attempting to reduce the number of SAEs required. For example, Layer Group SAE \cite{ghilardi2024efficient}, which we discussed previously, clusters multiple layers into layer groups and trains a single SAE per group instead of per layer. The emergence of such strategies demonstrates the significant computational burden imposed by layer-wise SAE training and the ongoing efforts to optimize it.




\begin{table*}[ht]
\centering
\caption{Evaluation of SAEs}
\label{tab:evaluate_sae}
\scalebox{0.9}{
\begin{tabular}{llccccccc}
\toprule
& & LLaMa Scope & Pythia SAE & \multicolumn{5}{c}{Gemma Scope} \\ \hline
& & & & L0:22  & L0:41  & L0:82  & L0:176 & L0:445  \\ \hline
Structural & Sparsity & 869.318 & 112.888 & 22.141 & 41.422 & 80.472 & 174.74 & 472.199 \\
        & MSE & 4.9E-5 & 0.015 & 2.125 & 1.836  & 1.539  & 1.203  & 0.707   \\
        & CE Loss & 1.00 & 0.940 & 0.974 & 0.984  & 0.988  & 0.993  & 0.998   \\
        & KL Div & 0.898 & -1 & 0.975  & 0.984  & 0.990  & 0.994  & 0.997   \\
        & Variance & 0.863 & 0.918 & 0.824  & 0.848  & 0.875  & 0.902  & 0.941   \\ \hline
Absorption  & Mean & 2.8E-3 & 0.227 & 0.287  & 0.267  & 0.105  & 0.055  & 0.1347  \\
        & Full & 4.7E-4 & 0.199 & 0.333  & 0.275  & 0.091  & 0.038  & 0.045   \\
SCR & Top 5 & 0.137 & 0.330  & 0.206  & 0.217  & 0.210  & 0.184  & 0.243   \\
        & Top 50  & 0.713 & 0.414 & 0.376  & 0.385  & 0.407  & 0.417  & 0.384   \\
        & Top 500 & -0.727 & 0.232 & 0.316  & 0.309  & 0.359  & 0.339  & 0.384   \\
Sparse Probing & LLM & 0.904 & 0.922 & \multicolumn{5}{c}{0.958} \\
        & SAE & 0.885 & 0.929 & 0.952  & 0.955  & 0.955  & 0.957  & 0.958   \\
\bottomrule
\end{tabular}}
\end{table*}

\section{Evaluation and Comparison of SAEs}
\label{sec:evaluate_sae}
As shown in Table \ref{tab:evaluate_sae}, we evaluate three series of SAEs: LLaMa Scope, Pythia SAE, and Gemma Scope. For LLaMa Scope, we use the llama-scope-lxa-8x SAE trained on layer 12 of the LLaMa 3.1 8B model. For Pythia, we evaluate the sae-bench-pythia70m-sweep-standard-ctx128-0712 SAE, trained on layer 5 of the Pythia 70M model. Within the Gemma Scope, we conduct an internal comparison across five gemma-scope-2b-pt-res SAEs, all trained on layer 12 of the Gemma2 2B model. These five SAEs differ only in their training L0 sparsity settings, meaning they were trained to activate different numbers of latents per input.

We report both structural and functional evaluation metrics. The structural metrics, listed at the top, include L0 Sparsity, Mean Squared Error (MSE), Cross-Entropy Loss, KL Divergence, and Explained Variance. The functional metrics, shown below, including Absorption, Spurious Correlation Removal (SCR), and Sparse Probing. In the Absorption metric, mean absorption captures the fraction of cases where the correct SAE latents fail to activate for a known feature, while another latent with similar semantics activates instead. Full absorption refers to the stricter case where none of the correct latents activate and the feature is entirely absorbed by an unintended latent. In SCR, we assess how well the SAE reduces unwanted correlations. The Top-5, Top-50, and Top-500 SCR scores measure how much debiasing can be achieved by removing top 5, 50, 500 latents respectively. Finally, Sparse Probing evaluates whether the SAE has learned interpretable and disentangled concepts. We compare the performance of sparse probes using SAE activations against a baseline probe using the LLM’s original dense activations. Higher score indicates that the SAE has successfully learned concept-specific features.

Shown in Table \ref{tab:evaluate_sae}, the LLaMa Scope SAE shows significantly higher L0 sparsity (869.3) compared to the others. The Gemma Scope SAEs span a range of sparsities, from highly sparse (L0:22 with 22.1 active latents) to relatively dense (L0:445 with 472.2). Within the Gemma Scope series, we observe that as sparsity denser, the MSE also decreases, indicating improved reconstruction accuracy. Similarly, both the cross-entropy loss and KL divergence scores increase toward 1, suggesting that denser SAEs preserve the model’s original predictive behavior more. The explained variance also improves with denser sparsity, rising from 0.824 at L0:22 to 0.941 at L0:445, which shows that denser SAEs are better at capturing the variance of the original activations.

In functional metrics, LLaMa Scope exhibits extremely low mean and full absorption values (0.0028 and 0.00047), likely due to its high redundancy from overactivation. Among the Gemma Scope models, L0:176 achieves the lowest mean (0.055) and full (0.038) absorption, suggesting a good trade-off between sparsity and disentanglement. Later, absorption increases again at L0:445, indicating that overly dense SAEs may begin to recombine multiple concepts into the same latent. Looking at SCR, all SAEs' performance improves with top-k increase. However, SCR scores begin to decrease at Top-500, indicating that over-ablation may unintentionally remove useful features. For Sparse Probing, we can see that across all models, SAE probes perform nearly as well as LLM-based probes, with accuracies consistently above 0.95 for Gemma Scope. Notably, the Pythia SAE and Gemma Scope L0:445 achieve sparse probing accuracies that even exceed the LLM baseline, suggesting that these models successfully capture clean, task-relevant features using a small set of latents.

\section{Other Applications of SAEs in LLMs}
\label{sec:applications_llm}

\subsection{Model Training}
SAEs are trained to obtain more sparse and interpretable features. The learned concepts and sparsity are both beneficial in model transparency, which can be utilized in model training to align the model with human understanding and improve model performance. Since SAEs include feature-level constraints,~\citet{yin2024direct} leverage these constraints to enable sparsity-enforced alignment in post-training. Their experiments demonstrate that this approach achieves superior performance across benchmark datasets with reduced computational costs. Similarly, combine learned concepts with next-token prediction training to build more transparent models. Specifically, they extract influential concepts on outputs from SAEs, then incorporate these concept vectors into hidden states by modifying token embeddings. Results show that models trained with this method perform better and exhibit greater robustness on token prediction and knowledge distillation across benchmark datasets~\cite{tack2025llm}. Moreover, SAEs' ability to provide large-scale explanations has been well explored. By examining the diversity of activated features, ~\citet{yang2025diversity} developed a new approach to augment data diversity. Another work uses task-specific features learned in SAEs to mitigate unintended features within models, significantly improving model generalization on real-world tasks~\cite{wu2025self}.

\section{SAE and Probing-Based Methods}
\label{sec:sae_probing}
In addition to SAEs, probing-based methods have emerged as a prominent family of techniques for interpreting and steering the internal representations of LLMs~\cite{giulianelli2018under, vig2020investigating, meng2022locating, guerner2023geometric, geiger2023causal}. These methods operate under the assumption that meaningful concepts are linearly separable within the model’s representation space~\cite{mikolov2013linguistic, pennington2014glove, park2023linear, nanda2023emergent}. Probing formulates steering as a supervised learning problem, which given a small set of labeled examples, the method learns low-rank projection vectors that can both detect and influence the presence of target concepts during generation. 


Based on the recent research, SAEs show certain limitations compared to linear probes for language model interpretation. First, the AXBENCH study~\cite{wu2025axbenchsteeringllmssimple}  demonstrates that even simple linear probe baselines like difference-in-means consistently outperform SAEs on concept detection tasks, with SAEs falling behind on both concept detection and model steering benchmarks. Second, Kantamneni et al.'s comprehensive evaluation~\cite{kantamneni2025sparseautoencodersusefulcase} across 113 probing datasets reveals that SAEs fail to provide consistent advantages in challenging regimes like data scarcity, class imbalance, label noise, and covariate shift. 
There remains a long way for the SAE research community to go in making sparse autoencoders more robust, generalizable, and useful in practical applications before they can reliably outperform simpler interpretability approaches.

\section{Research Challenges}
\label{sec:research_challenge}
In this section, we outline several critical research challenges with SAEs. Although SAEs have emerged as promising tools for providing large-scale, fine-grained interpretable explanations, these challenges could threaten the faithfulness, effectiveness, and efficiency of their applications.

\subsection{Incomplete Concept Dictionary}
SAEs are trained on large corpora of data encompassing various concepts. However, achieving comprehensive concept coverage remains challenging~\citep{muhamed2024decoding}. Additionally, the learning process of SAEs functions as a black box where learned concepts cannot be predetermined. Consequently, controlling the completeness of input and output concepts is nearly impossible.
Furthermore, explanations provided by SAEs may be incomplete or misleading due to the conceptual gaps. This limitation can result in unreliable interpretations when applying SAEs to complex reasoning tasks that require comprehensive knowledge representation.

\subsection{Lack Theoretical Foundations}
The development of SAEs is indeed based on assumptions of superposition and linear concept representation. Empirically, we've found it effective to construct high-level features through linear combinations of low-level features. However, our understanding of how these concepts are represented in hidden spaces and their spatial relationships remains limited. This limitation explains why we must derive combination parameters empirically rather than mathematically.
The validity and advancement of SAEs may remain unclear until we can properly demonstrate the correctness of these fundamental assumptions about concept representation and superposition in neural networks.

\subsection{Reconstruction Errors}
SAEs are trained by minimizing the reconstruction errors between original and reconstructed activations. However, these errors persist and remain poorly understood. Recent research by~\cite{gao2025scaling} demonstrates that reconstruction errors can produce significant performance degradation comparable to using a model with only 10\% of the pretraining compute. This finding raises substantial concerns about SAE accuracy as interpretability tools. Furthermore, the impact of these reconstruction errors on model generations has not been adequately measured. The field lacks output-centric metrics that could precisely quantify how reconstructed activations affect a model's final outputs. To advance our understanding of SAEs and their reliability as interpretability tools, developing metrics that directly measure the effect of reconstruction errors on generated content is essential.

\subsection{Computational Burden}
SAEs operate at the layer level for each model, mapping original activations to a much higher-dimensional representation space before reconstructing them back to the original space. This architecture necessitates that SAE parameters for a specific layer significantly outnumber the parameters of that original layer itself. Consequently, the overall training computation exceeds that of the original model training, particularly problematic for LLMs with billions of parameters.
The extensive computational resources required create a substantial barrier for researchers interested in investigating these methods. Furthermore, SAEs exhibit limited transferability across models, they must be trained specifically for each model and each layer, exacerbating the computational burden. This layer-specific and model-specific training requirement multiplies the already significant resource demands, further restricting accessibility for the broader research community.

\end{document}